\documentclass[10pt,journal,compsoc]{IEEEtran}
\usepackage{graphicx}
\usepackage{amsmath}
\usepackage{amssymb}

\usepackage{microtype}
\usepackage{color}
\usepackage{colortbl}
\usepackage{multirow}
\usepackage{makecell}
\usepackage{hhline}
\usepackage{cite}
\usepackage[american]{babel}
\usepackage{bbding}
\usepackage{cuted}
\usepackage{ragged2e}
\definecolor{lightgray}{gray}{.92}
\definecolor{tinygray}{gray}{.96}

\newcommand{\lxm}[1]{\textcolor[rgb]{0,0,0}{#1}}

\newcommand{\lxmday}[1]{\textcolor[rgb]{0,0,0}{#1}}
\newcommand{\lxmrb}[1]{\textcolor[rgb]{0,0,0}{#1}}

\newcommand{\etal}{\textit{et al}.}
\newcommand{\ie}{\textit{i}.\textit{e}.}
\newcommand{\eg}{\textit{e}.\textit{g}.}
\newcommand{\etc}{\textit{etc}}

\ifCLASSINFOpdf

\else

\fi

\usepackage[bookmarks,colorlinks,linkcolor=red, anchorcolor=blue, citecolor=green]{hyperref}
\begin{document}
	%
	\title{Learning Dual Memory Dictionaries for \\Blind Face Restoration}

	\author{Xiaoming Li,
		Shiguang Zhang,
		Shangchen Zhou,
		Lei Zhang,~\IEEEmembership{Fellow,~IEEE,}\\
		and Wangmeng Zuo,~\IEEEmembership{Senior Member,~IEEE,}
		\IEEEcompsocitemizethanks{\IEEEcompsocthanksitem X.  Li  is  with  the  Faculty of Computing,  Harbin Institute of Technology, Harbin, China, and also with the Department of Computing, the Hong Kong Polytechnic University, Hong Kong. (E-mail: csxmli@gmail.com)
			\IEEEcompsocthanksitem S. Zhang is with the Faculty of Computing, Harbin Institute of Technology, Harbin, China. (E-mail: zhangshiguang98@gmail.com)
			\IEEEcompsocthanksitem S. Zhou is with the School of Computer Science and Engineering, Nanyang Technological University, Singapore. (E-mail: shangchenzhou@gmail.com)
			\IEEEcompsocthanksitem L. Zhang is with the Department of Computing, the Hong Kong Polytechnic University, Hong Kong. (E-mail: cslzhang@comp.polyu.edu.hk)
			\IEEEcompsocthanksitem W. Zuo is with the Faculty of Computing, Harbin Institute of Technology, Harbin, China, and also with the Peng Cheng Lab, Shenzhen, Guangdong, China. (Corresponding author, E-mail: cswmzuo@gmail.com)
		}
		\thanks{Manuscript received xxx; revised xxx.}
	}


	\IEEEtitleabstractindextext{%
		\begin{abstract}
			Blind face restoration is a challenging task due to the unknown, unsynthesizable and complex degradation, yet is valuable in many practical applications. 
			To improve the performance of blind face restoration, recent works mainly treat the two aspects, \ie, \textbf{generic} and \textbf{specific} restoration, separately. 
			\lxm{In particular, \textbf{generic} restoration attempts to restore the results through general facial structure prior, while on the one hand, cannot generalize to real-world degraded observations due to the limited capability of direct CNNs' mappings in learning blind restoration, and on the other hand, fails to exploit the identity-specific {details}. 
				On the contrary, \textbf{specific} restoration aims to incorporate the identity features from the reference of the same identity, in which the requirement of proper reference severely limits the application scenarios. }
			\lxm{Generally, it is a challenging and intractable task to improve the photo-realistic performance of blind restoration and adaptively handle the generic and specific restoration scenarios with a single unified model.}
			\lxm{Instead of implicitly learning the mapping from a low-quality image to its high-quality counterpart, this paper suggests a DMDNet by explicitly memorizing the generic and specific features through dual dictionaries.}
			First, the generic dictionary learns the general facial priors from high-quality images of any identity, while the specific dictionary stores the identity-belonging features for each person individually. 
			Second, to handle the degraded input with or without specific reference, dictionary transform module is suggested to %
			read the relevant details from the dual dictionaries which are subsequently fused into the input features.
			Finally, multi-scale dictionaries are leveraged to benefit the coarse-to-fine restoration. 
			The whole framework \lxm{including the generic and specific dictionaries is optimized in an end-to-end manner} and can be \lxm{flexibly} plugged into different application scenarios. 
			\lxm{Moreover, a new high-quality dataset, termed CelebRef-HQ, is constructed to promote the exploration of specific face restoration in the high-resolution space.}
			Experimental results demonstrate that the proposed DMDNet performs favorably against the state of the arts in both quantitative and qualitative evaluation, and generates more photo-realistic results on the real-world low-quality images. 
			The codes, models {and the CelebRef-HQ dataset} will be publicly available at \url{https://github.com/csxmli2016/DMDNet}.
		\end{abstract}
		
		}

	\maketitle

	\IEEEdisplaynontitleabstractindextext
	
	\IEEEpeerreviewmaketitle

	\IEEEraisesectionheading{\section{Introduction}\label{sec:introduction}}
	\IEEEPARstart{T}he task of blind face restoration is to recover high-quality textures from real-world degraded observations, without knowing the degradation types or parameters. Thus, this task has promising potential values in many practical applications (\eg, faces in albums, films, old photos, \etc), yet suffers from intractable challenges due to the complex, unknown and unsynthesizable degradation. 
	{A natural solution is to utilize the plain convolutional networks (CNNs) to directly learn the mapping from a low-quality (LQ) image to its high-quality (HQ) counterpart. 
		However, the difficulties in synthesizing realistic LQ images for constituting the training pairs and the limited capability of such plain frameworks make it fail to handle the blind restoration task.
	}
	Existing deep restoration networks, \ie, image super-resolution~\cite{dong2014learning,wang2018esrgan,zhang2018rcan}, denoising~\cite{zhang2017beyond,zhang2018ffdnet}, deblurring~\cite{kupyn2018deblurgan, Kupyn_2019_ICCV}, or compression artifact removal~\cite{dong2015compression}, generally perform limited in restoring real-world natural images. However, as for some special categories, \eg, face images, considerable progress has been made and two {main} kinds of methods have been proposed for blind face restoration, \ie, generic and specific restoration.

	{On the one hand, many works exploit the facial structure priors (facial landmarks, parsing maps, heatmaps, \etc) to improve the restoration performance~\cite{Chen_2018_CVPR,bulat2018super,progressive_face_sr,chen2020progressive,Yu_2018_ECCV}. }
	However, all these priors encapsulate semantic label or geometry information rather than small-scale textures, thereby bringing limited improvements for the restoration of fine details.
	{Besides, the synthetic training pairs easily make them overfit to these types of known degradation and fail to handle the real-world LQ images with unseen degradation.}
	{These~make them inappropriate to handle the blind restoration task.}
	{Moreover, all these generated textures are learned from the general facial prior, which inevitably ignores the individual details, especially when the input suffers from severe degradation. }
	{On the other hand, to improve the restoration performance and generalization ability, specific restoration is further suggested by utilizing the existing HQ images as references~\cite{li2018learning,dogan2019exemplar,Li_2020_CVPR}. With the development of image capturing and sharing technologies, these types of HQ images for each person usually are available, \eg, albums in smartphones, celebrities in multimedia. }
	{Although the external HQ references could diminish the sensitivity of degradation types and incorporate the identity-belonging details, all these methods have limited application scenarios due to the requirement of the HQ references which should have the same identity, similar poses and expressions}.
	\lxm{However, such proper references are usually inaccessible in many cases, making them hardly work in most application scenarios.}

	{Inspired by the benefits that the incorporation of HQ images in specific restoration brings to the blind restoration, our previous work DFDNet~\cite{Li_2020_ECCV} proposed to exploit component dictionary for storing the {general} HQ textures of eyes, nose, and mouth. 
		Though significant improvements have been achieved for blind restoration, it still ignores the identity-belonging details when the degradation is severe. 
		In this paper, we further extend this work with DMDNet by learning the dual (generic and specific) dictionaries to memorize the general and individual features, respectively.  
	}
	\lxmrb{Intuitively, each person shares similar structures with other identities, but meanwhile, has unique and specific characteristics.
		In particular, the generic dictionary in our DMDNet learns the general facial prior from large amounts of HQ images from different identities, while the specific one stores the individual features from the corresponding HQ references of the same identity. So the generic dictionary in our DMDNet aims to provide general guidance for improving the generalization ability of blind face restoration, but may fail to preserve the identity-related textures when handling the severely degraded images that have unrecognizable appearances. In contrast, the specific dictionary could compensate for the generic dictionary by providing the identity-related textures, which contributes to better generation of identity-related textures. Such specific dictionary is also practically available, \eg, smartphones that support to group the face images according to their identities, and famous actors that have many HQ portraits on the Internet. By using these existing HQ references from the same identity, it would be valuable in personalized restoration in smartphones or actor face restoration in old films.}
	\lxm{In contrast to plain CNNs in learning the restoration mappings, the external dictionaries in our DMDNet can eliminate the difficulties of learning blind restoration through memorizing the desired HQ features, and more importantly, can guide the generation of photo-realistic and identity-aware details for the final restoration results.}
	\lxm{Furthermore, both of the dual dictionaries can benefit and compensate for each other, \ie, the specific one can provide and enhance the identity-belonging details, while the generic one can provide abundant priors when the optimal reference is not available or cannot satisfy the consistency of poses and expressions. }
	\lxm{By separately learning the generic and specific features through dual dictionaries, our DMDNet can flexibly restore the degraded observation to the general one or the specific one under different application scenarios.}

	With the learned dual memory dictionaries, we subsequently investigate the manner of how to utilize them to guide the blind restoration. Among these specific restoration methods, GFRNet~\cite{li2018learning} and GWAINet~\cite{dogan2019exemplar} only adopt a single HQ reference. Though ASFFNet~\cite{Li_2020_CVPR} exploits the multi-exemplar in this task, it also selects only one exemplar that has the optimal poses for the final restoration process. These methods cannot fully utilize the reference features because the unaligned poses and expressions cannot provide accurate guidance, further resulting in performance degradation. To handle this problem, we propose the dictionary transform module by taking all the dictionary features into the reconstruction process. First, each feature in the dictionary contains two parts, \ie, value and key, in which the value is expected to contain the HQ structures that are beneficial for the reconstruction quality, and the key is proposed to match with the degraded input. Second, by computing the similarity of the keys from the dictionaries and the queries from the LQ input, the final read values are obtained by adaptively fusing the generic and specific features. This can not only utilize both the generic and specific priors, {but also eliminate the matching gap between the LQ input and HQ dictionaries}. Besides, the specific dictionary is easy to extend by adding more diverse HQ exemplars individually, which is flexible to plug into the model without fine-tuning. 
	Third, to achieve the coarse-to-fine restoration, the dictionary transform module is performed in multi-scale feature spaces.

	{Benefited from the high-resolution face datasets, \eg, CelebA-HQ~\cite{karras2017progressive} and FFHQ~\cite{karras2019style}, generic restoration methods can generate high-resolution results (\ie, $\geq$ 512$\times$512). 
		However, specific restoration methods usually are tested on $256\times256$ images because existing individual datasets which have the corresponding references \lxm{and initially are proposed for face or attribute recognition,} \eg, CelebA~\cite{liu2015faceattributes}, VggFace2~\cite{cao2018vggface2}, CASIA-WebFace~\cite{yi2014learning}, do not have HQ images for training  (\ie, $\geq$ 512$\times$512), thereby limiting their practical applicability.}
	To address this issue, we build a new dataset, called CelebRef-HQ, by crawling the high-resolution images of recent celebrities. It mainly contains 1,005 identities that cover different ages, ethnicities, expressions, \etc. Each person has 2$\sim$21 HQ references, resulting in a total of 10,555 images.

	Extensive experiments are conducted to evaluate the effectiveness of our proposed DMDNet in different scenarios. The quantitative and qualitative results on synthetic and real-world LQ images show that our DMDNet performs favorably against the state of the arts in generating plausible and photo-realistic results, and moreover, exhibits great generalization ability in handling complex and unknown degradation.

	\lxm{This paper is a substantial extension of our previous work DFDNet~\cite{Li_2020_ECCV}. In comparison to DFDNet, we extend the single general dictionary to the dual dictionaries. Moreover, both the dual dictionaries and the whole restoration network can be optimized by the reconstruction loss. Finally, the dictionary transform module is introduced to adaptively read and fuse the generic and specific features, making our DMDNet flexibly handle the degraded inputs with or without references in a unified framework. Meanwhile, a new HQ dataset, namely CelebRef-HQ, is proposed to facilitate the research of high-resolution specific face restoration.
	}
	The main contributions of this work are summarized as follows:
	\begin{itemize}
		\vspace{-3pt}
		\setlength{\parsep}{0pt}
		\item \lxm{Instead of directly learning the restoration mapping, we explicitly memorize the external HQ features to facilitate the learning of blind face restoration.}
		\item \lxm{To handle the degraded observations with or without references in a unified framework, we propose the DMDNet by learning the dual (\ie, generic and specific) memory dictionaries to store the general and individual high-quality features separately.}
		\item \lxm{For combining the dual memory dictionaries which have inconsistent item numbers, a dictionary transform module is introduced for adaptive fusion. 
			This can well utilize the complementary details of these two dictionaries, and more importantly, can also perform favorably when the references are not available.}
		\item {We construct a new face dataset with high-quality references, named CelebRef-HQ, to promote the specific restoration task on handling high-resolution images.}
		\item \lxm{Experimental results demonstrate that our DMDNet outperforms the state-of-the-arts  in generating favorable and photo-realistic results, and shows great generalization ability in different application scenarios.}
	\end{itemize}
	
	\lxm{The remainder of this paper is organized as follows. In Section~\ref{sec:2},  relevant studies including generic and specific face restoration are reviewed. Section~\ref{sec:3} presents the details of learning the dual dictionaries and dictionary transform module. Section~\ref{sec:4} reports the experimental results and analyses. Finally, concluding remarks are provided in Section~\ref{sec:5}.}
	
	\section{Related Work}\label{sec:2}
	In this section, we briefly review the relevant methods from two aspects, \ie,  generic and specific face restoration.
	
	\subsection{Generic Face Restoration}
	
	Image restoration has achieved remarkable success in many tasks, \ie, single image super-resolution~\cite{dong2014learning,kim2016accurate,ledig2017photo,zhang2018rcan,Zhang_2019_CVPR,dai2019second,zhang2020deep,zhou2020cross}, denoising~\cite{zhang2017beyond,zhang2018ffdnet,yang2017bm3d,guo2019toward,DBSN}, deblurring~\cite{nah2017deep,kupyn2018deblurgan,zhang2019deep,Zhang_2020_CVPR}, and compression artifact removal~\cite{dong2015compression,galteri2017deep,guo2017one,Xu_2019_ICCV}, due to the rapid development of convolutional neural networks (CNNs) and the generative adversarial networks (GANs)~\cite{goodfellow2014generative}. Face images that can be regarded as a subspace of natural manifold, have received considerable attention because of its special structures and wide application scenarios~\cite{zhu2016deep,cao2017attention,huang2017wavelet,xu2017learning,chrysos2017deep,Yu_2018_ECCV,Yu_2018_CVPR,progressive_face_sr,Chen_2018_CVPR,chen2020progressive,Li_2020_ECCV,wang2021towards}. 
	Zhu \etal~\cite{zhu2016deep} proposed a bi-network to learn face hallucination and dense correspondence estimation in a unified framework. 
	Cao \etal~\cite{cao2017attention} utilized reinforcement learning to discover the attended patches and then exploit the global interdependency of the image to enhance the facial part. 
	Yu \etal~conducted a series of works by using GANs to ultra-resolve face images~\cite{yu2016ultra} and extended it to handle unaligned and noisy faces~\cite{yu2017hallucinating}.
	Huang \etal~\cite{huang2017wavelet} suggested predicting the wavelet coefficients to reconstruct the restoration results. 
	Subsequently, to improve the restoration performance, many works attempt to incorporate facial prior into the degraded input~\cite{pan2014deblurring,shen2018deep,Yu_2018_ECCV,Chen_2018_CVPR}.
	\lxm{Pan \etal~\cite{pan2014deblurring} proposed a maximum posteriori (MAP) deblurring algorithm from an exemplar dataset to express the facial structure information. Then}
	both Chen \etal~\cite{Chen_2018_CVPR} and Adrian \etal~\cite{bulat2018super} used the facial landmark heatmaps to embed the geometry prior. Similarly, Yu \etal~\cite{Yu_2018_ECCV} suggested predicting the heatmaps of facial components for incorporating structural information. Shen \etal~\cite{shen2018deep} exploited to impose local structure by learning global semantic labels. 
	\lxmrb{Zhang~\etal~introduced the internal and recursive external copy operation from the well-illuminated facial details for the joint face hallucination and illumination compensation tasks~\cite{recpgan}, and proposed the facial component prior for the joint face hallucination and frontalization tasks~\cite{vividgan}.}
	To reduce the ambiguity in face super-resolution, Yu \etal~\cite{Yu_2018_CVPR} incorporated the face attribute vector into the LR feature space. 
	%
	%
	\lxm{Chen \etal~\cite{chen2020progressive} reformulated the blind restoration task as semantic-aware style transformation, in which it takes the degraded input to generate the semantic style details to guide the image generation. With the powerful generation ability of pre-trained StyleGANs~\cite{karras2019style,karras2020analyzing}, Wang~\etal~\cite{wang2021towards}, Chan~\etal~\cite{chan2020glean} and Yang~\etal~\cite{Yang2021GPEN} adopted them as generative face prior to improve and stabilize the restoration quality. 
	\lxmrb{Most recently, vector-quantization based methods (\ie, RestoreFormer~\cite{wang2022restoreformer} and VQFR~\cite{gu2022vqfr}) have been proposed to learn the dictionary and codebook from HQ face images, which is utilized to restore the LQ input by reconstructing the whole results, including background, hair, \etc.}
	It can be seen that all the aforementioned methods adopt the general face prior that is learned from a large amount of training data. Though the restoration results are pleasing, the generated details are hard to guarantee that they belong to the same identity, especially handling the severely degraded~images. Here, we call these types of methods as generic face restoration.}

	\subsection{Specific Face Restoration}
	Different from the above methods, the specific face restoration (also known as reference-based face restoration) tends to incorporate the identity-belonging details to the degraded input. In their seminal work, 
	\lxmday{Li \etal~\cite{li2018learning} modified this problem by taking a frontal HQ image to enhance the LQ input of the same identity. To eliminate the inconsistent poses and expressions, they adopted a WarpNet to deform the guidance to the desired one which was utilized in the following reconstruction.}
	\lxmday{Subsequently, Dogan \etal~\cite{dogan2019exemplar} proposed a GWANet to learn the warped guidance without requiring facial landmarks.
		After that, Li \etal~\cite{Li_2020_CVPR} extended this task from the single guidance to multiple exemplars.} To reduce the effect of misalignment, they firstly selected the optimal exemplar with similar poses and expressions with the degraded input.  Then, feature and spatial alignment, as well as the adaptive feature fusion are proposed to generate the final results. Albeit the HQ features of the same identity have been incorporated into the LQ input, we analyze that 1) both methods~\cite{li2018learning,dogan2019exemplar,Li_2020_CVPR} take only one image in the restoration process. When the selected guidance is not available or is hard to satisfy the requirement of poses, it is hard to benefit the reconstruction, resulting in limited improvement. 2) It is intractable for users to collect the exemplars that cover nearly all the poses and expressions, otherwise, the misalignment easily degrades the performance~\cite{Li_2020_CVPR}. All these motivate us to exploit the combination of both generic and specific face restoration by adaptively utilizing their merits.
	
	\section{Methodology}\label{sec:3}
	
	\begin{figure*}[t]
		\centering
		\includegraphics[width=1.00\textwidth]{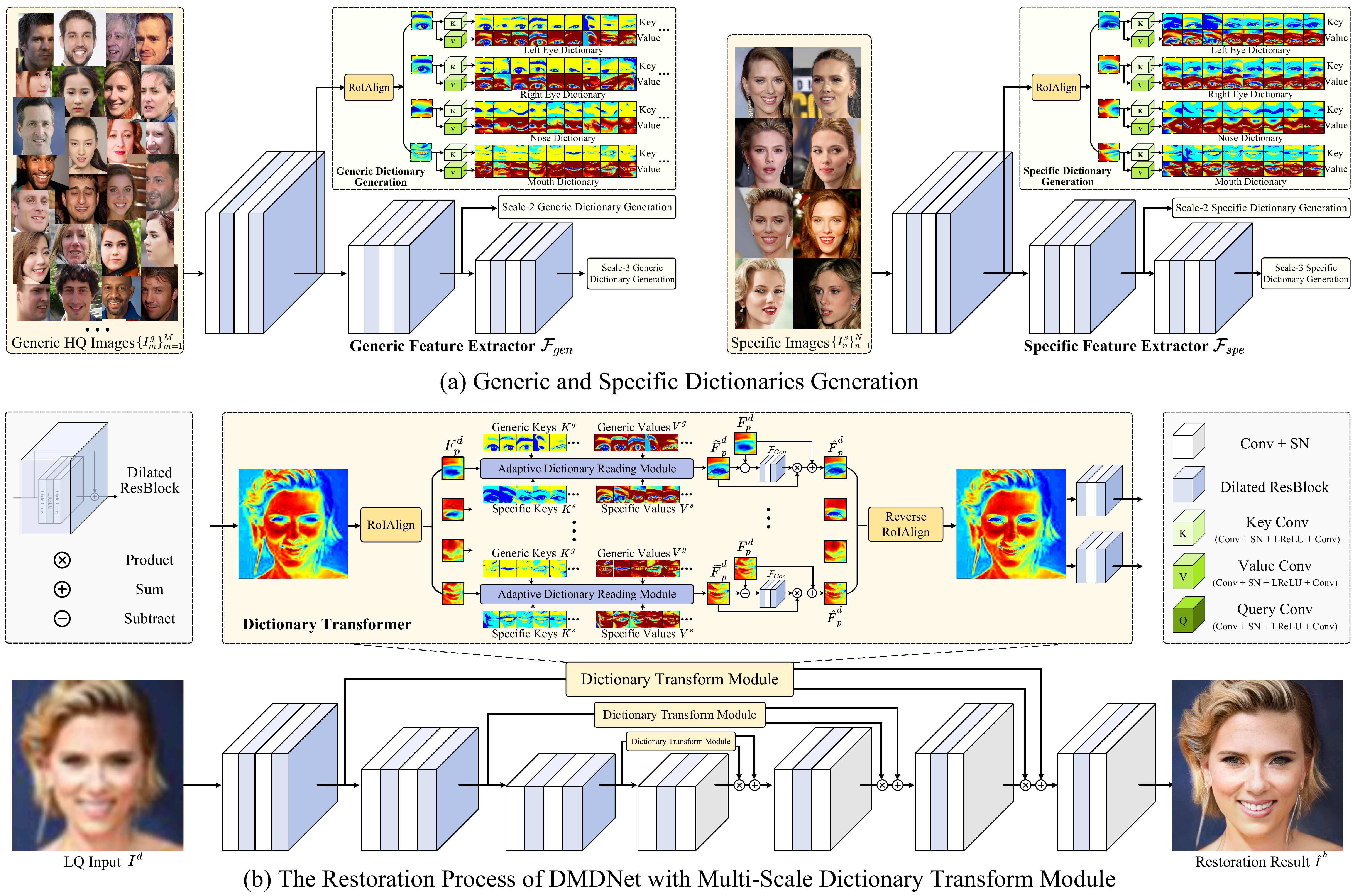}
		\caption{Overview of our proposed DMDNet. For each component feature that is obtained through RoIAlign operation, the generic dictionary is conducted from the HQ images with different identities, while the specific one is generated from the HQ references with the same identity. 
		The dictionary transform module is proposed to adaptively read and fuse the generic and specific dictionary features.
		Confidence score is finally adopted to handle the input with different degradation levels.
		}
		\label{fig:pipeline}
	\end{figure*}

	Given a degraded LQ input $I^d$ that suffers from unknown degradation, our DMDNet aims to generate a photo-realistic result $\hat{I}^h$ by using the dual memory dictionaries. The proposed blind face restoration model can be defined as:
	\begin{equation}
		\label{eqn:whole}
		\hat{I}^h = \mathcal{F}\left( I^d|Dic^g, Dic^s;\Theta \right),
	\end{equation}
	\lxm{where $Dic^g$ and $Dic^s$ are the generic and specific dictionaries, respectively, $\Theta$ denotes the learnable model parameters. In particular, $Dic^g$ and $Dic^s$ are constructed from the sets of generic HQ images $\{I^g_m\}^M_{m=1}$ with different identities and specific HQ images $\{I^s_n\}^N_{n=1}$ with the same identity, respectively.}
	\lxm{The whole framework is illustrated in Figure~\ref{fig:pipeline}. It mainly contains three parts, \ie, the generic and specific feature extractors in (\textcolor{red}{a}), and the restoration process by utilizing the dual dictionaries in (\textcolor{red}{b}). All three parts are incorporated in a unified framework. In the following, we first describe the generation of the dual dictionaries. Then, the dictionary transform module is introduced to flexibly and adaptively guide the restoration process. Finally, we give the learning objective for training the whole framework.} 

	\subsection{Generic Dictionary Generation}
	\lxm{Without loss of generality, the dictionary is usually used to store data in the format of \{$key: value$\} pairs. Similarly, in this work, the value is expected to contain the desired HQ features which will benefit the later texture enhancement, while the key is adopted to work as an index for matching with the LQ input. Although our pioneer work DFDNet~\cite{Li_2020_ECCV} also generates the general facial dictionary, 
		we analyze that it has the following issues that may degrade the performance. 
		First, the dictionary in DFDNet is conducted off-line by extracting features from the pre-trained VGGFace~\cite{cao2018vggface2}, {so both the encoder and the dictionary} cannot be end-to-end optimized by the final reconstruction loss, which may cause performance degradation. Second, the dictionary in DFDNet only contains value item (without key), thus the feature matching between the LQ input and the HQ values in the dictionary is sometimes not accurate, resulting in  worse guidance for restoration. 
		To address the optimization issue and bridge the matching gap, in this paper we suggest an end-to-end mechanism to conduct the generic key-value dictionary $Dic^g$ through the following three stages.}
	
	\noindent{\textbf{Initialization.}}  The generic feature extractor $\mathcal{F}_{gen}$ is proposed to extract the features of generic HQ images to conduct the initial dictionary. Here we suppose that there are $Y$ items for each component (\ie, eyes, nose, mouth) in the generic dictionary. This generation process can be formulated as:
	{
		\begin{equation}
			Dic^{g} =\! \{key^g_y, value^g_y\}^Y_{y=1} =\! \{\mathcal{F}_{gen}(I^g_y;\Theta_{gen})\}^Y_{y=1},
		\end{equation} 
		where $\{I^g_y\}^Y_{y=1}$ are the \lxmday{$Y$} images that are randomly selected from {generic} HQ images with different identities, $\Theta_{gen}$ is the learnable parameters of generic feature extractor. By using the $Dic^g$ in the later dictionary transform module, $\mathcal{F}_{gen}$ can be optimized by minimizing the {objective loss (Eqn.~\ref{eqn:final})}.}
	
	\noindent{\textbf{Forward Update.}} After the training loss tends to be stable, we take the latest $Dic^g$ as the initial generic dictionary. 
	{We note that the selected $Y$ (128 in this work) images may not be enough to cover all the poses and expressions.
		Following \cite{kaiser2017learning,Zhu_2020_CVPR}, we further update the key-values by considering other images for generalizing to different poses and expressions,}
	\begin{equation}
		\begin{split}
			value^{*} &= \gamma_{v} \cdot value + (1-\gamma_v)\cdot value^{gt}, \\
			key^{*} &=  \gamma_{k} \cdot key + (1-\gamma_k)\cdot key^{gt}, 
		\end{split}
	\end{equation}
	{where $value$ and $key$ are from the initial $Dic^g$, $value^{gt}$ and $key^{gt}$ are generated from $\mathcal{F}_{gen}(I^{gt};\Theta_{gen})$, $\gamma_k$ and $\gamma_v$ are the learnable parameters. $I^{gt}$ is the ground-truth of LQ input.}
	
	\noindent{\textbf{Backward Update.}} \lxmrb{Before this stage, the $key$ and $value$ are the intermediate features that are obtained from the convolution layers. In this stage, we remove the generic feature extractor and transfer the $key$ and $value$ in the former optimized dictionary $Dic^g$ as learnable parameters with a small learning rate $\eta$ (\ie, $2\times10^{-6}$), which are directly optimized through the gradients from the final learning objective $\mathcal{L}$ in Eqn.~\ref{eqn:final}. This is formulated as:}
	\begin{equation}
		\setlength{\abovedisplayskip}{5pt}
		\setlength{\belowdisplayskip}{5pt}
		\begin{split}
			{value}^{*} &=  {value} - \eta \cdot \frac{\partial \mathcal{L}}{{\partial value}}, \\
			key^{*} &=  key - \eta \cdot \frac{\partial \mathcal{L}}{\partial key}, 
		\end{split}
	\end{equation}
	\lxmrb{Then the module in Eqn.~\ref{eqn:whole} can be reformulated as:}
	\begin{equation}
		\label{eqn:whole2}
		\hat{I}^h = \mathcal{F}\left( I^d|\{I^s_n\}^N_{n=1};\Theta, \Theta_{Dic^g}\right),
	\end{equation}
	\lxmrb{where $\Theta$ and $\Theta_{Dic^g}$ are the network parameters and generic dictionary, respectively. %
		$\{I^s_n\}^N_{n=1}$ is the sets of specific HQ images. After the losses on the validation set become stable, the generic $key$ and $value$ are further transferred as model buffer that are not optimized for efficiently fine-tuning $\Theta$.}

	\subsection{Specific Dictionary Generation}
	We observe that each identity usually has uncertain numbers of HQ references, \eg, some people like famous celebrities may have abundant HQ images, while the others may only have very limited ones.
	So it is hard and unfair to maintain the same number of dictionary items to balance this situation.
	In this work, we design a dynamic manner to conduct the specific dictionary. Given the HQ references $I^s$, the generation of specific dictionary $Dic^s$ is formulated as:
	\begin{equation}
		Dic^s \!= \{key^s_n, value^s_n\}^N_{n=1} \!=\{\mathcal{F}_{spe}(I^s_n; \Theta_{spe})\}^N_{n=1},
	\end{equation}
	where $\mathcal{F}_{spe}$ is the specific feature extractor, $\Theta_{spe}$ is the learnable parameter and is initialized by $\Theta_{gen}$. $N$ is the number of the specific HQ images for the current given input $I^d$ ($N \in \{0:21\}$). It should be noted that  each identity may have different $N$ in the specific dictionary but {share} the same {$Y$ ($Y=128$)} in the generic dictionary. Instead of taking the $Dic^g$ as learnable parameters in conducting the generic dictionary, we optimize $\Theta_{spe}$ in the whole training process and generate the $Dic^s$ dynamically. \lxmrb{Thus, the specific dictionary is the intermediate features obtained from convolution layers during the whole training phase. This allows us to  flexibly add new HQ references into the specific dictionary when individual HQ images are available.}

	\begin{figure}[t]
		\centering
		\includegraphics[width=.5\textwidth]{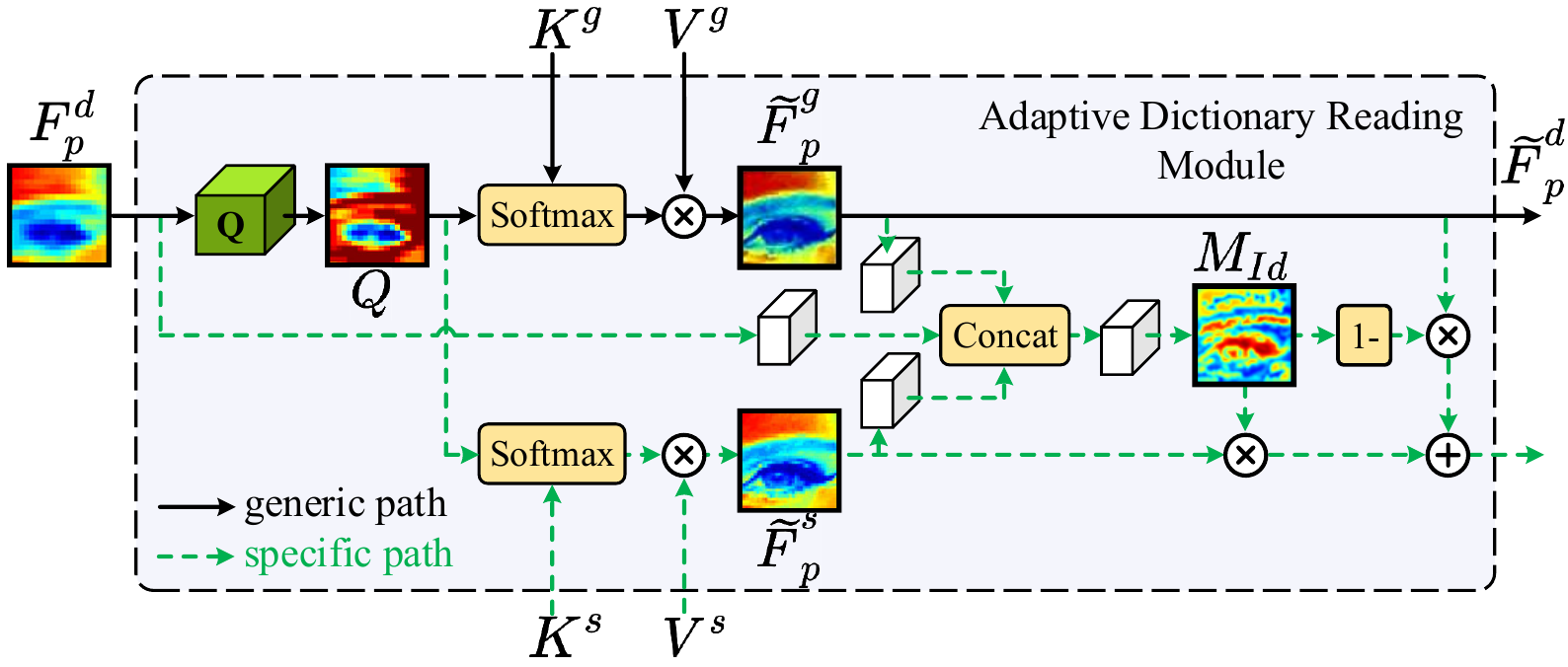}
		\caption{Details of the adaptive dictionary reading module.
		}
		\label{fig:ddrm}
	\end{figure}
	
	\subsection{Dictionary Transform Module}
	After generating the dual memory dictionaries, \ie, $Dic^g$ and $Dic^s$, the Dictionary transform module is then proposed to combine them to reconstruct the HQ results. 
	Inspired by the attention structure in transformer~\cite{vaswani2017attention}, which can handle the unconstrained sequences, 
	we can also utilize it to fuse the inconsistent dictionary items.
	\lxmrb{Albeit each LQ input may have different numbers of dual dictionary items (\ie, {$Y+N$}), the attention mechanism can circumvent this problem by computing the similarities between keys and queries, which are further performed on the values. 
		Denote by $F^d_p$ the component features that are cropped from the LQ input with RoIAlign operation and $p \in \{\text{left eye, right eye, nose and mouth}\}$.
		As shown in Figure~\ref{fig:ddrm}, the corresponding $queries$ for each component are generated from $F^d_p$ through several convolution blocks. With the $query$ from the LQ input and the $\{key:value\}$ from the former conducted dual dictionaries, the reading process for each dictionary is formulated as:}
	\begin{equation}
		\widetilde{F}^\star_p = softmax\left(\frac{Q \cdot K^\star}{\sqrt{d}}\right)V^\star,
	\end{equation}
	\lxmrb{where $Q$, $K^\star$ and $V^\star$ are the sets of $query$, $key^\star$ and $value^\star$, respectively. Here, $^\star$ represents the generic or specific dictionary.  
		$d$ is introduced to stabilize the gradient and is set to 64 in our experimental settings.
		When the specific dictionary is not available, the output $\widetilde{F}^d_p$ of dictionary reading module  is equal to $\widetilde{F}^g_p$. Otherwise, the output is:}
	\begin{equation}
		\label{eqn:adaread}
		\widetilde{F}^d_p = M_{\textit{Id}} \cdot \widetilde{F}^s_p + (1-M_{\textit{Id}})\cdot \widetilde{F}^g_p\,,
	\end{equation}
	\lxmrb{where $M_{\textit{Id}}$ is obtained by taking the LQ feature $F^d_p$, generic result $\widetilde{F}^g_p$ and specific result $\widetilde{F}^s_p$ to predict the identity score which is used to adaptively incorporate the identity-related features into the final results.}
	
	\lxm{To adaptively fuse the read features $\widetilde{F}^d_p$ to the LQ features $F^d_p$ which may have different degradation levels, we follow our former DFDNet~\cite{Li_2020_ECCV} by computing the confidence score from the residual between $\widetilde{F}^d_p$ and $F^d_p$. This is defined as:}
	\begin{equation}
		\hat{F}^d_p = {F}^d_p + \widetilde{F}^d_p\cdot\mathcal{F}_{Con}(\widetilde{F}^d_p - F^d_p;\Theta_{Con}),
	\end{equation}
	\lxm{where $\Theta_{Con}$ is the learnable parameters with two convolutional layers. The final fused component features $\hat{F}^d_p$ is then {copy-and-pasted} to the original LQ features through reverse RoIAlign operation.
		Following \cite{chen2020progressive,wang2021towards}, we adopt the spatial feature transform (SFT)~\cite{wang2018recovering} by learning the  feature modulation function to incorporate the fused features upon the decoder features. This affine modulation is defined as: }
	\begin{equation}
		F^{out} = \alpha \cdot F^{decoder} + \beta,
	\end{equation}
	{where $F^{decoder}$ is the output of the previous decoder module.  The scale $\alpha$ and shift $\beta$ parameters are both element-wise weights which have the same size with $F^{decoder}$.}
	
	To benefit the coarse-to-fine restoration, we adopt multiple-scale dictionary transform modules in different feature spaces. The whole framework of our DMDNet is a modified UNet~\cite{ronneberger2015u}, in which the dictionary transform module is utilized in each skip connection. 

	\subsection{Learning Objective}
	To train our DMDNet, two types of loss functions are collaborated together, \ie, reconstruction loss and photo-realistic loss. In general, the former one is introduced to constrain the results $\hat{I}^h$ close to the ground-truth $I^{gt}$, while the latter one is utilized to encourage $\hat{I}^h$ within the natural HQ manifold and meanwhile recover the realistic details.
	
	\noindent\textbf{Reconstruction Loss}. We adopt the commonly used mean square error loss (MSE) to minimize the difference between the result $\hat{I}^h$ and its corresponding ground-truth $I^{gt}$ in both pixel and feature space. The pixel loss is defined as:
	\begin{equation}
		\mathcal{L}_{mse}=\frac{1}{CHW}\left\|\hat{I}^h-I^{gt}\right\|^{2},
	\end{equation}
	where $C$, $H$, $W$ are the channel, height, and width of the image, respectively. Following perceptual loss \cite{johnson2016perceptual}, the feature difference between $\hat{I}^h$ and $I^{gt}$ is computed by:
	\begin{equation}
		\mathcal{L}_{perc}= \sum_{i=1}^{4} \frac{1}{\mathcal{C}_{i} \mathcal{H}_{i} \mathcal{W}_{i}}\left\|\Phi_{i}(\hat{I}^h)-\Phi_{i}(I^{gt})\right\|^{2},
	\end{equation}
	where $\mathcal{C}_i$, $\mathcal{H}_i$, and $\mathcal{W}_i$ are the dimensions from the $i$-\textit{th} convolution layer of the pretrained VGG-19 model $\Phi$ \cite{simonyan2014very}. 
	
	The overall reconstruction loss is defined as:
	\begin{equation}
		\mathcal{L}_{rec} = \lambda_{mse}\mathcal{L}_{mse} + \lambda_{perc}\mathcal{L}_{perc},
	\end{equation}
	where the trade-off parameters $\lambda_{mse}$ and $\lambda_{perc}$ are set to 300 and 1 in our experimental settings, respectively.
	
	\noindent\textbf{Photo-realistic Loss}. Two loss terms are considered for the photo-realistic reconstruction. The first one is style loss~\cite{gatys2016image}, which is defined on the Gram matrix and shows great performance in generating visually plausible details~\cite{liu2018image,chen2020progressive,wang2021towards}. The style loss is defined as:
	\begin{equation}
		\mathcal{L}_{style} \!\!=\!\! \sum_{i=1}^{4}\! \frac{1}{\mathcal{C}_{i} \mathcal{H}_{i} \mathcal{W}_{i}} \!\left\|\Phi_{i}(\hat{I}^h)^T\!\Phi_{i}(\hat{I}^h)\!-\!\Phi_{i}(I^{gt})^T\!\Phi_{i}\!\left(I^{gt}\right)\right\|^{2},
	\end{equation}
	in which the variate has the same definition as $\mathcal{L}_{perc}$. 
	
	The second one is the commonly used adversarial loss for improving visual quality. Following \cite{Li_2020_ECCV,Li_2020_CVPR}, we adopt the SNGAN~\cite{miyato2018spectral} with multiple constraints in different image resolutions. The whole training objective for discriminator $D$ and generator $G$ are formulated as:
	\begin{equation}
		\begin{aligned}
			\mathcal{L}_{D}=&\sum_{r}^{R} \mathbb{E}_{I_{\downarrow r}^{gt} \sim P(I_{\downarrow r}^{gt})}\left[\min (0, D_{r}(I_{\downarrow r}^{gt})-1)\right]\\
			&+\mathbb{E}_{\hat{I}^h_{\downarrow r} \sim P(\hat{I}^h_{\downarrow r})}
			\left[\min (0,-1-D_{r}(\hat{I}^h_{\downarrow r}))\right],
		\end{aligned}
	\end{equation}
	\begin{equation}
		\mathcal{L}_{G}\!=\!-\lambda_{a,r}\!\sum_{r}^{R}\!\mathbb{E}_{I^{d} \sim P(I^{d})}\!\left[D_r(\mathcal{F}(I^{d}|\{I^s_n\}^N_{n=1}; \Theta, \Theta_{Dic^g})_{\downarrow r})\right],
	\end{equation}
	where  ${\downarrow_r}$ is the down-sampling operation with scale factor $r$ and $r \in \{1,2,4\}$. $\lambda_{a,r}$ is set to $\{4, 1, 0.5\}$, respectively.
	
	The overall photo-realistic loss is defined as: 
	\begin{equation}
		\mathcal{L}_{real} = \lambda_{style}\mathcal{L}_{style}+\mathcal{L}_G,
	\end{equation}
	where $\lambda_{style}$ is the trade-off parameter and is set to 0.1.
	
	\lxm{By taking both the reconstruction loss and the photo-realistic loss, the final learning objective is formulated as: }
	\begin{equation}
		\label{eqn:final}
		\mathcal{L} = \mathcal{L}_{rec} + \mathcal{L}_{real}.
	\end{equation}
	
	\begin{table*}[ht]
		\caption{\lxm{Quantitative comparison of generic restoration methods on three test datasets (\ie, FFHQ, CelebA-HQ and CelebRef-HQ) and on $\times$4 and $\times$8 tasks. Here, $\uparrow$ ($\downarrow$) indicates higher (lower) is better.} The best two results are highlighted in \textbf{bold} and \underline{underline}, respectively. }
		\centering
		\footnotesize
		\renewcommand\arraystretch{1.7}
		%
		\setlength{\tabcolsep}{0.238mm}
		{	
			\begin{tabular}{|c| c c c c| c c c c| c c c c| c c c c| c c c c c|  c c c c c|}
				\hline
				\rowcolor{lightgray}
				& \multicolumn{8}{c|}{\textbf{FFHQ~\cite{karras2019style}}} & \multicolumn{8}{c|}{\textbf{CelebA-HQ~\cite{karras2017progressive}}}&
				\multicolumn{10}{c|}{\textbf{CelebRef-HQ}} \\[0pt]
				\hhline{>{\arrayrulecolor{lightgray}}-|>{\arrayrulecolor{black}}--------------------------}
				\rowcolor{lightgray}& \multicolumn{4}{c|}{$\times 4$} & \multicolumn{4}{c|}{$\times 8$} & \multicolumn{4}{c|}{$\times 4$} & \multicolumn{4}{c|}{$\times 8$}&
				\multicolumn{5}{c|}{$\times 4$} & \multicolumn{5}{c|}{$\times 8$} \\[-4pt]
				\rowcolor{lightgray}\multirow{-2.7}{*}{\makecell[c]{\textbf{Methods}}}&
				\tiny PSNR$\uparrow$ &\tiny SSIM$\uparrow$ & \tiny LPIPS$\downarrow$ &\tiny FID$\downarrow$ & \tiny PSNR$\uparrow$ &\tiny SSIM$\uparrow$ & \tiny LPIPS$\downarrow$ &\tiny FID$\downarrow$ & \tiny PSNR$\uparrow$ &\tiny SSIM$\uparrow$ & \tiny LPIPS$\downarrow$ &\tiny FID$\downarrow$ & \tiny PSNR$\uparrow$ &\tiny SSIM$\uparrow$ & \tiny LPIPS$\downarrow$ &\tiny FID$\downarrow$ & \tiny PSNR$\uparrow$ &\tiny SSIM$\uparrow$ & \tiny LPIPS$\downarrow$ &\tiny FID$\downarrow$ & \tiny Id$\uparrow$& \tiny PSNR$\uparrow$ &\tiny SSIM$\uparrow$ & \tiny LPIPS$\downarrow$ &\tiny FID$\downarrow$ & \tiny Id$\uparrow$\\[-1pt]
				\hline \hline
				PULSE  & 21.55   & .760   & .413  & 94.71  & 21.40  & .757   & .420  & 78.46
				& 22.21  & .783  & .399  & 86.93  & 22.09  & .780  & .405  & 89.98 
				& 22.44  & .795  & .353  & 80.37 & .324 & 22.36  & .793  & .361  & 85.10  & .291 \\
				HiFace   & \underline{27.75}   & \underline{.881}   & {.203}  & 16.55 & 25.31  & .836   &  .303  & 32.53
				& \underline{27.48}  & \underline{.875}  & {.228}  & {14.29}  & \underline{25.49}  & \underline{.835}  & .313   & 25.73 
				& \underline{28.40}  & \underline{.895}  & {.180}  & 16.72 & \underline{.729} & 25.97  & .846  & .269  & 31.59  & \textbf{.549} \\
				DFDNet   & 25.94   & .832   & .218  & {14.43} & 24.40  & .792   & .288  &\underline{20.62}
				& 25.92  & .822  & .234  & {14.44}  & 24.44  & .783  & .296  & {17.79} 
				& 26.44  & .834  & .183 & \textbf{9.75} & .702 & 25.04  & .798  & .242  & {14.72}  & .515\\
				PSFRGAN   & 27.46   & .878   & .211  & 16.36  & \underline{25.62}  & \underline{.844}   &  {.266}   &21.73
				& 25.83  & .855  & .241  & 16.75  & 24.92  & .833  & {.286} & 20.16 
				& 28.05  & .883  & .184 & 11.20  & \textbf{.730} & \underline{26.52}  & \underline{.854}  &  {.229} & 15.52 & \underline{.542} \\
				GPEN & 27.64 & .879 & \underline{.191} & 13.49
				& 25.24 & .832 & .239 & 20.65
				& 27.34 & .873 & \underline{.211} & \underline{14.26}
				& 25.42 & .834 & .258 & 17.68
				& 28.31 & \underline{.895} & .179 & 10.35 & .712
				& 25.77 & .841 & .210 & 14.71 & .503\\
				GFPGAN &  27.40 &  .876 & {.192} & \underline{13.45}
				& 25.26 & .832 & \underline{.232} & {20.64}
				& 27.29& .871&{.213} &\textbf{14.23} 
				& 25.38&.833 &\underline{.256} &\underline{17.63}
				& 28.27&.894 & \underline{.172} & 10.33 & .689
				& 25.76&.841 & \underline{.207}&\underline{14.69} & .455\\
				\hline
				Ours$^g$  & \textbf{28.32}  &  \textbf{.896}  & \textbf{.174}  & \textbf{13.16}  & \textbf{26.27}  &  \textbf{.859}  & \textbf{.221} & \textbf{19.10}
				& \textbf{28.17} & \textbf{.891} & \textbf{.197}  & 14.53 & \textbf{26.26}  & \textbf{.851}  & \textbf{.244}  & \textbf{17.55}
				& \textbf{28.91}  & \textbf{.901}  & \textbf{.169}  & \underline{10.31} & .715 & \textbf{27.03}  & \textbf{.870}  & \textbf{.201}  &  \textbf{14.56} & .504\\
				\hline
		\end{tabular}}
		\label{tab:generic}
	\end{table*}

	\section{Experiments}\label{sec:4}
	Since the proposed DMDNet can handle both generic and specific face restorations on degraded inputs with random combinations of several common degradation, in this work, we mainly compare these two types of competing methods on restoring high-resolution (\ie, $512\times512$) images. \lxmrb{For more general comparison, we select five state-of-the-art methods (\ie, HiFaceGAN~\cite{yang2020hifacegan}, PULSE~\cite{menon2020pulse},  PSFRGAN~\cite{chen2020progressive}, GPEN~\cite{Yang2021GPEN}, GFPGAN~\cite{wang2021towards}) and our pioneer work DFDNet}. In terms of specific restoration, we compare our DMDNet with single (GFRNet~\cite{li2018learning}) and multiple (ASFFNet~\cite{Li_2020_CVPR}) exemplar-based methods. Since GFRNet~\cite{li2018learning} and ASFFNet~\cite{Li_2020_CVPR} can only handle $256\times256$ images due to the limited training data, we retrain them with our CelebRef-HQ dataset for a fair comparison, which are denoted as GFRNet* and ASFFNet* in the later subsections. Following \cite{Li_2020_ECCV,Li_2020_CVPR}, we also synthesize the \lxm{test} LQ input with random conjunction of noise, blurring and JPEG compression on $\times4$ and $\times8$ tasks for quantitative evaluation. 
	Furthermore, we also give the qualitative comparison on both synthetic and real-world degraded images to assess the abilities of generating photo-realistic details and generalizing to different degradation.
	
	\subsection{CelebRef-HQ Dataset}
	To exploit the specific face image restoration in high-resolution space, we {collect} the CelebRef-HQ dataset by crawling the celebrities of recent years from Bing Images~\footnote{https://www.bing.com/images}. Each image should first satisfy that the face resolution is at least $512\times512$. Then Laplacian gradient is utilized to assess the image quality and remove all but those with high scores. Finally, to exclude the outlier samples that are not from the same identity, we adopt Arcface~\cite{deng2019arcface} to compute the identity distance for  each person until the recognition accuracy achieves 100\%. We also manually check each person to guarantee the visual quality and identity. To sum up, our CelebRef-HQ contains 1,005 identities, which has a total of 10,555 images and covers different ages, genders, ethnicities, backgrounds, poses, expressions, \etc. After that, we adopt \cite{bulat2017far} to detect their 68 facial landmarks and utilize them to crop and align the raw image to a fixed size, \ie, $512\times512$. Furthermore, we divide it into three parts,
	\ie, a training set of 805, a validation set of 50, and a test set of 150. The sets are not overlapped in terms of either identity and image. 
	\lxm{The CelebRef-HQ dataset will be publicly available to benefit reference-based high-quality face restoration.}%
	
	\subsection{Datasets and Implementation Details}
	\lxm{For generic face restoration, all the competing methods (\ie, HiFaceGAN~\cite{yang2020hifacegan}, PULSE~\cite{menon2020pulse},  PSFRGAN~\cite{chen2020progressive}, GPEN~\cite{Yang2021GPEN} and GFPGAN~\cite{wang2021towards}) adopt FFHQ dataset~\cite{karras2019style} for training}, we also utilize it in our experiments for training and evaluating the generic restoration as well as learning the generic memory dictionary. In particular, we randomly select 66,000 images for training, 2,000 for validation, and the remaining 2,000 for testing. Besides, we also build another test set from CelebA-HQ~\cite{karras2017progressive} by randomly selecting 2,000 images.
	\lxm{The test set of our CelebRef-HQ is adopted to evaluate both the generic and specific restoration.
		PSNR, SSIM~\cite{wang2004image}, LPIPS~\cite{zhang2018perceptual}, and FID~\cite{heusel2017gans} are reported to assess the quantitative performance.}
	
	Following the competing methods \cite{li2018learning,yang2020hifacegan,wang2021towards,Yang2021GPEN,chen2020progressive,Li_2020_CVPR}, we adopt the same degradation model by considering the random combinations of common degradation, \ie, blur, noise, down-sampling and JPEG compression to synthesize the training and testing pairs, formulated as:
	\begin{equation}
		I^{d}=\left((I^{gt} \otimes \mathbf{k}_{\varrho})_{\downarrow_{r}}+\mathbf{n}_{\sigma}\right)_{\mathbf{c}_{q}}.
	\end{equation}
	Without loss of generality, $\mathbf{k}$, $\downarrow$, $\mathbf{n}$ and $\mathbf{c}$ denote the Gaussian blur kernel, down-sampling, Gaussian noise and JPEG compression, respectively. In particular, $\varrho \in \{1:0.1:3\}$, $r \in \{1:0.1:10\}$, $\sigma \in \{0:1:15\}$ and $q \in \{50:1:100\}$. 
	
	The whole model and experiments are conducted on a PC server with four Tesla V100 GPUs. ADAM optimizer~\cite{kingma2014adam} with $\beta_1=0.5$ and $\beta_2=0.999$ is utilized to train our 
	DMDNet.  The batch size is set to 8 and the initial learning rate for training the model parameters $\Theta$ is set to $2\times10^{-4}$ ($2\times10^{-6}$ for $\Theta_{Dic^g}$) and will decrease by 0.5 when the reconstruction loss on validation set tends to be stable. Data augmentation, \eg, horizontal flipping and color jittering~\cite{zoph2019learning}, are also exploited to increase image diversities.
	
	\subsection{Quantitative Evaluation}

	\begin{table}[t]
		\caption{\lxm{Quantitative comparison of specific restoration methods on CelebRef-HQ test set and on $\times$4 and $\times$8 tasks.}}
		\centering
		\renewcommand\arraystretch{1.5}
		\footnotesize
		\setlength{\tabcolsep}{0.25mm}
		{
			\begin{tabular}{|c| c c c c c| c c c c c|}
				\hline
				\rowcolor{lightgray}\rowcolor{lightgray}
				\rowcolor{lightgray}& \multicolumn{5}{c|}{$\times 4$} & \multicolumn{5}{c|}{$\times 8$} \\[-3pt]
				\rowcolor{lightgray}\multirow{-1.8}{*}{\makecell[c]{\textbf{Methods}}}&
				\scriptsize PSNR$\uparrow$ & \scriptsize SSIM$\uparrow$ &  \scriptsize LPIPS$\downarrow$ & \scriptsize FID$\downarrow$ &\scriptsize  Id$\uparrow$ & \scriptsize  PSNR$\uparrow$ & \scriptsize SSIM$\uparrow$ & \scriptsize  LPIPS$\downarrow$ & \scriptsize  FID$\downarrow$ &\scriptsize  Id$\uparrow$ \\
				\hline \hline
				GFRNet*
				& 27.59  & .878  & .203  & 19.86 & .733 & 26.21  & .850  & .243  &  19.32 & .625\\
				ASFFNet*   
				& 28.03  & .883  & .182  & 13.85 & \underline{.784} & 26.55  & .855  & .241  &  15.71 & \underline{.706}\\
				Ours$^g$
				&{28.91}  & {.901}  & {.169}  & {10.31} & .715 & {27.03}  & {.870}  & {.201}  &  {14.56}  & .504\\
				Ours (\textit{Full})& \textbf{28.97}  & \textbf{.902}  &  \textbf{.166} & \textbf{10.30}  & \textbf{.793} & \textbf{27.37}  & \textbf{.875}  & \textbf{.194}  & \textbf{14.43} & \textbf{.715} \\
				\hline
		\end{tabular}}
		\label{tab:specific}
	\end{table}
	
	\begin{figure*}[h]
		\setlength{\abovecaptionskip}{3pt}
		\centering
		\includegraphics[width=1.0\textwidth]{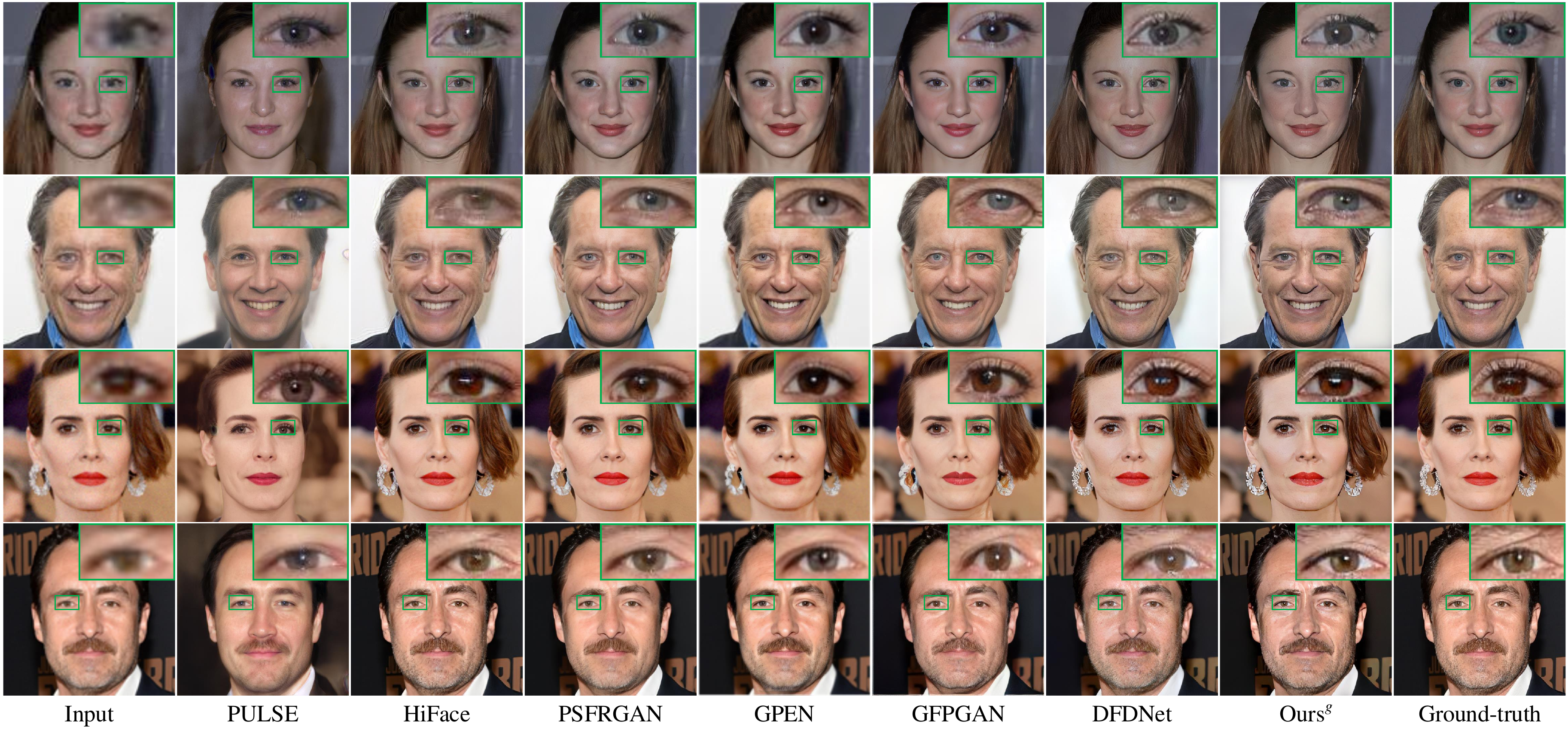}
		\caption{Visual comparison of generic  restoration methods on handling $\times4$  task. Best view it by zooming in the screen to see the details.
		}
		\vspace{-3pt}
		\label{fig:visualcompare_gx4}
	\end{figure*}
	
	\begin{figure*}[t]
		\setlength{\abovecaptionskip}{3pt}
		\centering
		\includegraphics[width=1.0\textwidth]{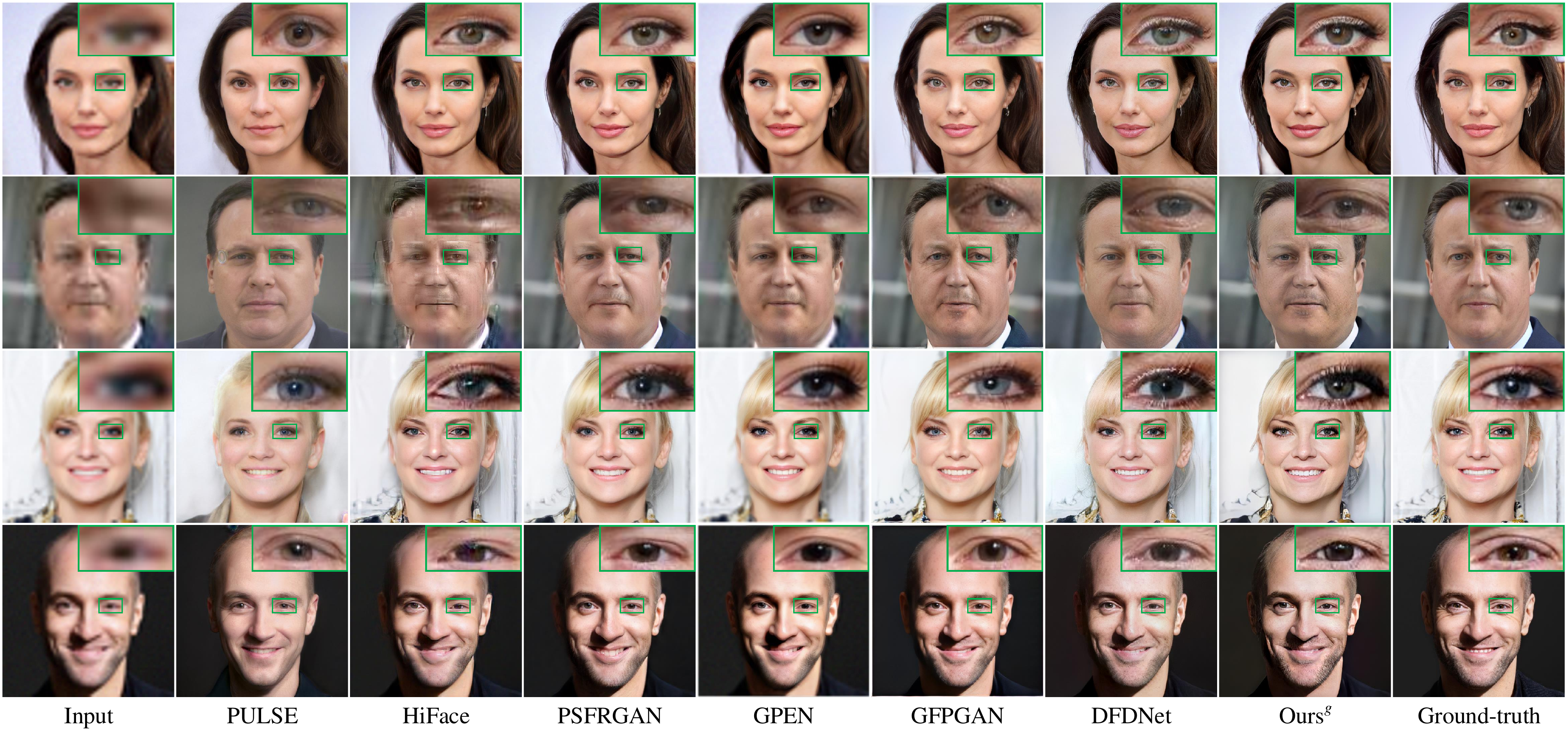}
		\caption{Visual comparison of generic restoration methods on handling $\times8$  task. Best view it by zooming in the screen to see the details.
		}
		\label{fig:visualcompare_gx8}
		\vspace{-3.6pt}
	\end{figure*}

	\lxm{Table~\ref{tab:generic} lists the quantitative results of $\times4$ and $\times8$ super-resolution tasks on three test datasets. The quantitative comparison is conducted on both generic and specific restoration tasks. As for the first one, we compare our DMDNet with the state-of-the-art face restoration methods, which cover different aspects, \ie, GAN prior based PULSE~\cite{menon2020pulse}, GPEN~\cite{Yang2021GPEN}, and GFPGAN~\cite{wang2021towards}, semantic map based PSFRGAN~\cite{chen2020progressive}, and our previous offline dictionary based DFDNet~\cite{Li_2020_ECCV}, \etc. }
	Except PULSE~\cite{menon2020pulse} which adopts the iterative optimization to minimize the distance between input and output, the remaining competing methods adopt the same training dataset and degradation model, so we can directly use their official model for a fair comparison.
	Ours$^g$ represents DMDNet with only generic dictionary. We can find that 1) our DMDNet achieves the best performance by a large margin in most metrics (\eg, 0.49 dB in $\times4$ and 0.65 dB in $\times8$  on average higher than the 2-\textit{nd} best method). 
	2) \lxm{Though our pioneer DFDNet attains the 
		comparable performance against others in LPIPS and FID, which is more consistent with human perception on visual quality, 
		it has inferior results on PSNR and SSIM. We analyze that this is mainly caused by its fixed encoder  and offline dictionary, which cannot be optimized for the reconstruction. 
		Similarly, GFPGAN~\cite{wang2021towards} and GPEN~\cite{Yang2021GPEN} achieved nearly the second best performance in LPIPS and FID metric, which show the powerful ability of GAN prior in generating photo-realistic results, but may fail to maintain their original structures that result in the relatively poor PSNR and SSIM.}
	Instead, the best performance of our DMDNet can be ascribed to the effectiveness of our trainable network and generic dictionary as well as the dictionary transform module. 
	3) Although all the methods are trained with FFHQ~\cite{karras2019style}, our DMDNet shows great generalization ability to other test sets.

	\begin{figure}[t]
		\centering
		\includegraphics[width=.49\textwidth]{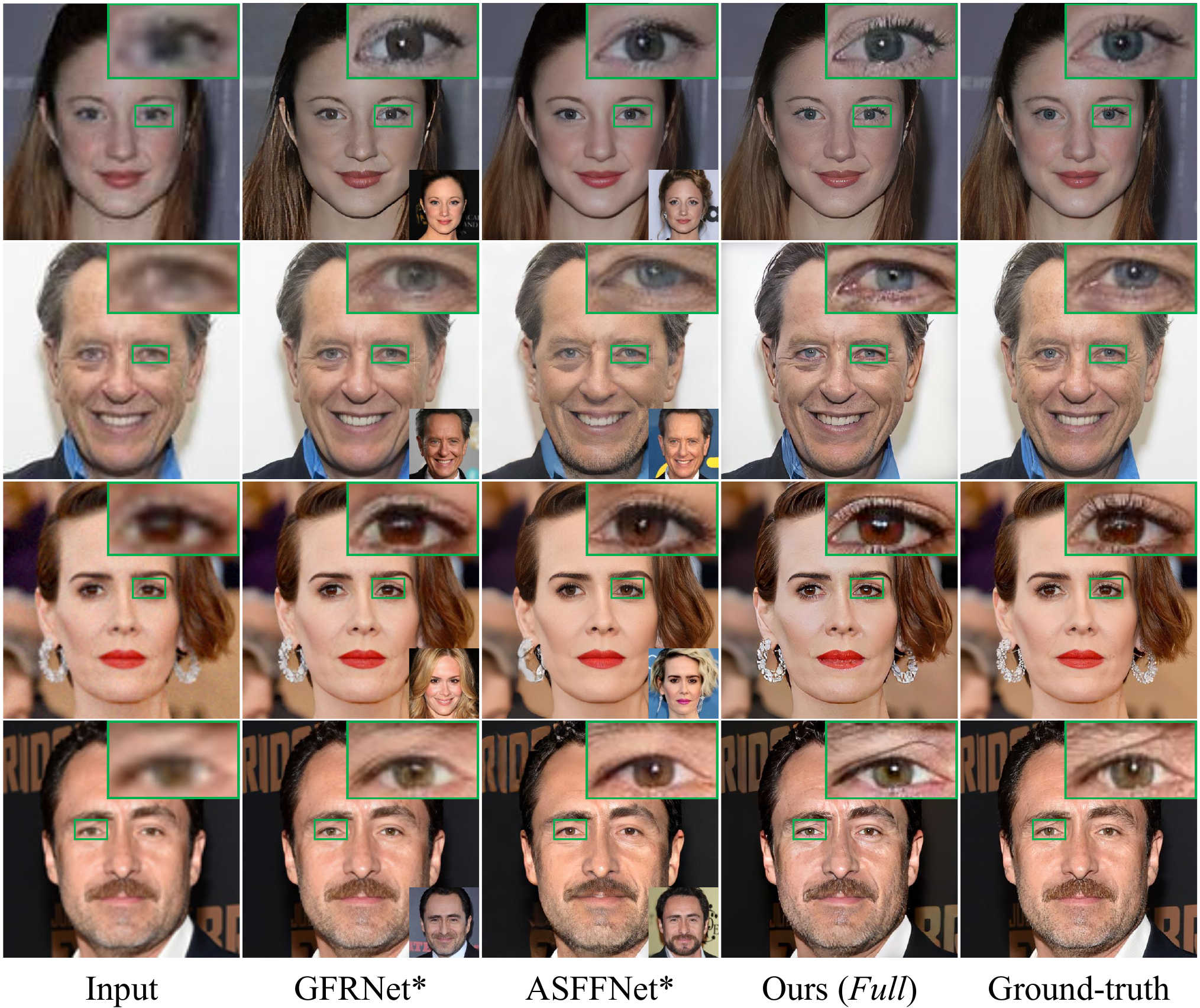}
		\caption{Visual comparison of specific restoration on $\times4$  task. Close-up in the right bottom is the selected reference for GFRNet* and ASFFNet*. 
		}
		\label{fig:visualcompare_sx4}
	\end{figure}

	In terms of specific restoration, both GFRNet~\cite{li2018learning} and ASFFNet~\cite{Li_2020_CVPR} are retrained with our CelebRef-HQ dataset for a fair comparison. 
	Following their experimental settings, {we select a frontal reference for each identity in GFRNet*, and all the references are adopted as the reference pool for ASFFNet*}. 
	The evaluations including $\times4$ and $\times8$ SR tasks are conducted on the CelebRef-HQ test set  and the quantitative performance is reported in Table~\ref{tab:specific}. 
	Ours {(\textit{Full}) represents} the proposed DMDNet by using the dual memory dictionaries. We can observe that, on the one hand, as for the moderate degradation (\ie, $\times4$), Ours  (\textit{Full}) performs on par with Ours$^g$, but has a obvious improvement on $\times8$ (\eg, 0.34 dB in $\times8$ \textit{vs} 0.06 dB in $\times4$), which indicates that the specific dictionary $Dic^s$ can incorporate the useful details for the final reconstruction when the degradation is severe. We analyze that these useful details should be the identity-belonging features that cannot be represented by the general facial priors, which further validate the effectiveness and necessity of our specific dictionary in face restoration task. On the other hand, both GFRNet* and ASFFNet* only adopt a single reference in their reconstruction process,  the inconsistency of poses and expressions between the LQ input and the single reference may explain their 
	\lxm{worse performance than our generic version. By taking both the generic and specific dictionaries in a unified framework, our DMDNet achieves the best performance, indicating the complementary effectiveness of generic prior may bring to the restoration.}

	\subsection{Qualitative Evaluation}
	
\begin{figure}[!t]
		\centering
		\includegraphics[width=.49\textwidth]{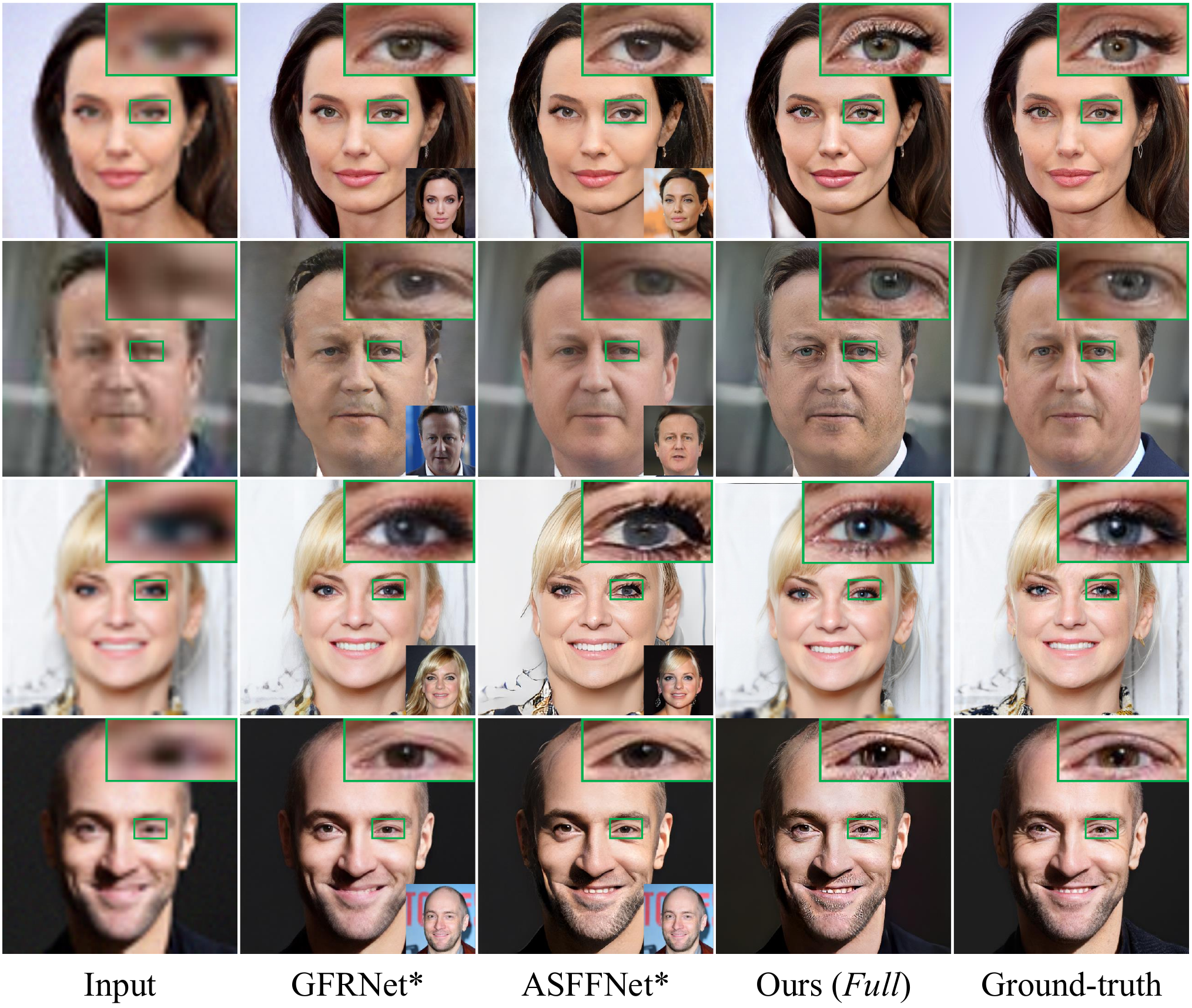}
		\caption{Visual comparison of specific restoration on $\times8$  task. Close-up in the right bottom is the selected reference for GFRNet* and ASFFNet*.
		}
		\label{fig:visualcompare_sx8}
	\end{figure}

	{Figures~\ref{fig:visualcompare_gx4}, \ref{fig:visualcompare_gx8} and Figures~\ref{fig:visualcompare_sx4}, \ref{fig:visualcompare_sx8} show the visual comparison with the competing methods on both generic  and specific restoration on handling $\times4$ and $\times8$ tasks, respectively. } From the restoration details we can have the following observations: 1) as for the generic restoration, all the competing methods can generate plausible results, but fail to produce identity-belonging details. 2) Though DFDNet conducts the component dictionaries, their performance in generating clearer structure \lxm{(\eg, eyelid)} is still inferior to Ours$^g$, indicating the effectiveness of our dictionary transform module by using the trainable Query-Key-Value mechanism. 3) As for the specific restoration, both GFRNet* and ASFFNet* have frontal and optimal HQ references, respectively. But the inconsistent poses and expressions may diminish the dependency on references, leading to their poor performance. 4) More importantly, all the competing methods including Ours$^g$ fail to generate the identity-belonging details, \eg, their pupil color and double-eyelids are easily changed (zoom in the close-up in the upper right to see the details).
	\lxmrb{Although the generative prior based GPEN~\cite{Yang2021GPEN} and GFPGAN~\cite{wang2021towards} show great ability in  generating plausible results, the difficulties in fitting the input with complex degradation to the exactly accurate $W$ space make it easily lose the original features (\ie, pupil color).
		Instead, with the complementary metrics of both generic and specific dictionaries, Ours (\textit{Full}) can generate  consistent details and structures with the ground-truth, indicating that the specific dictionary in our DMDNet can 
		learn the identity-belonging details, which will be utilized to enhance the individual textures in the restoration process. }

\begin{figure*}[t]
		\centering
		\vspace{-5pt}
		\includegraphics[width=1.0\textwidth]{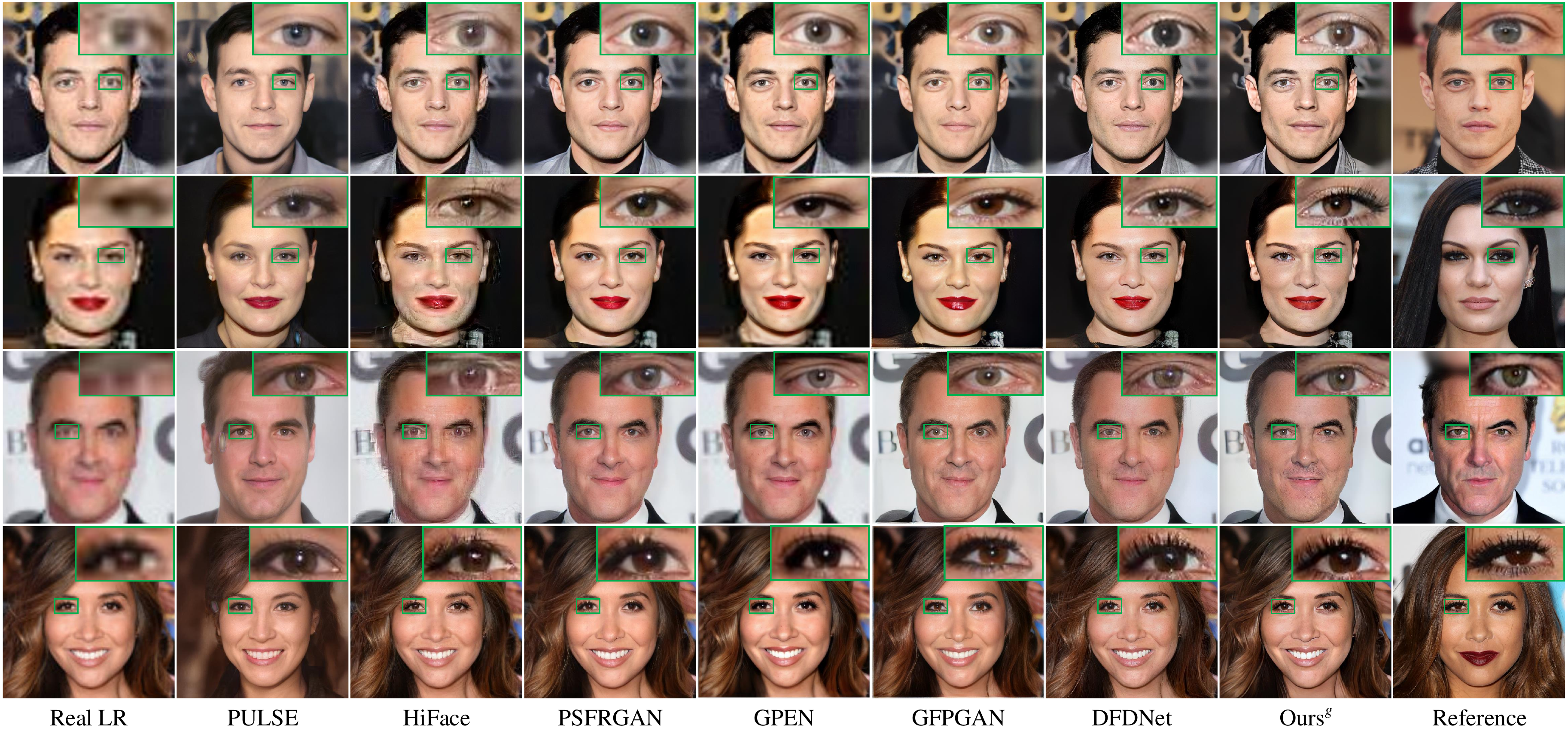}
		\caption{Visual comparison of generic restoration methods on handling real-world low-quality images. The last column is the selected reference from the same identity to show the identity-aware structures. Best view it by zooming in the screen to see the details.
		}
		\label{fig:realcompare}
		\vspace{-8pt}
	\end{figure*}

\begin{figure}[t]
		\setlength{\abovecaptionskip}{2pt}
		\centering
		\vspace{-5pt}
		\includegraphics[width=.49\textwidth]{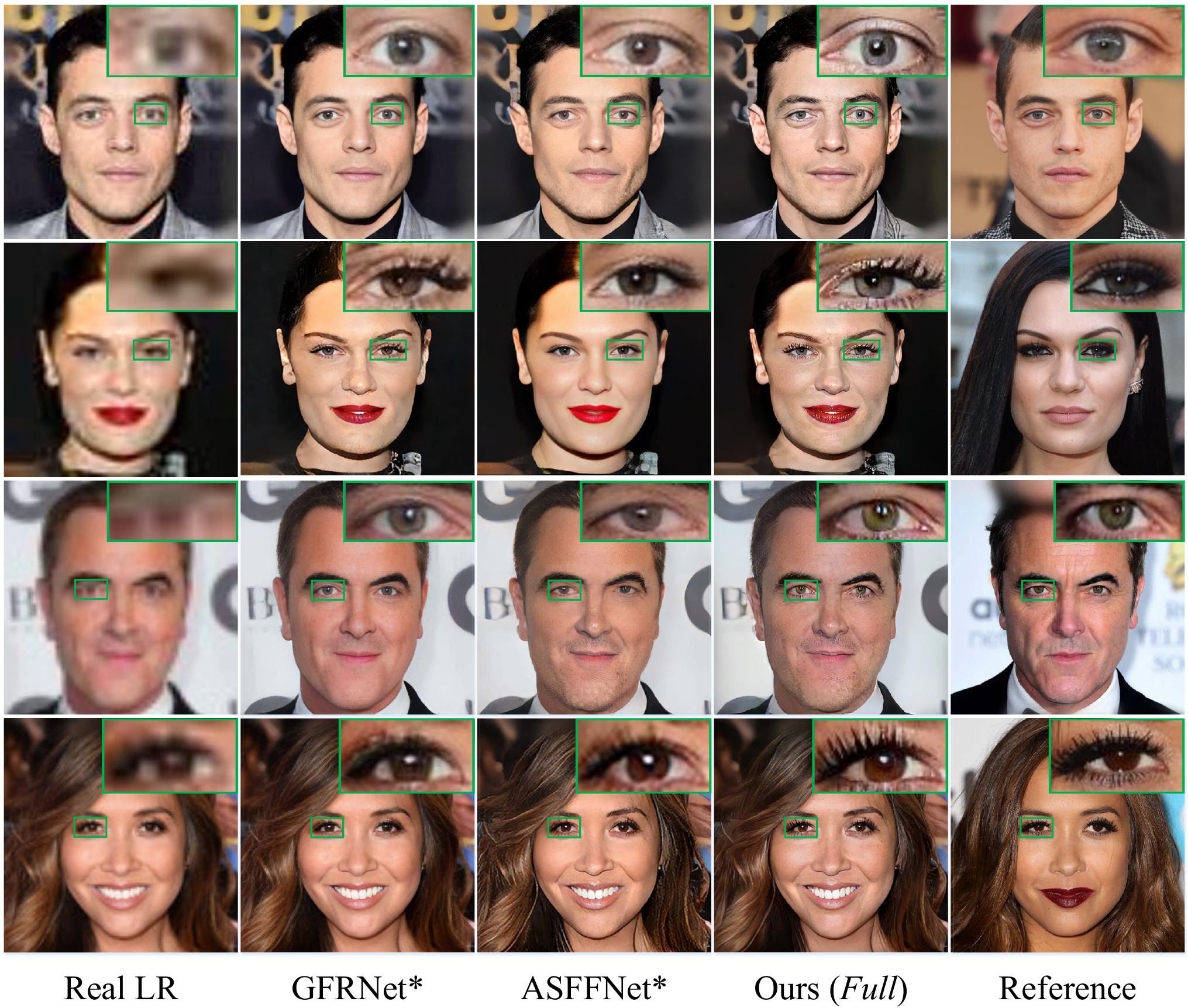}
		\caption{Visual comparison of specific restoration methods on handling real-world low-quality images. The last column is the selected reference from the same identity to show the identity-aware structures.
		}
		\vspace{-10pt}
		\label{fig:realcompare2}
	\end{figure}

	To further show the effectiveness of our DMDNet in handling the real-world degraded images,
	we conduct a CelebRef-LQ test set by crawling the face images with resolution lower than $80\times80$. Generic and specific restoration results are shown in Figures~\ref{fig:realcompare} and \ref{fig:realcompare2}, respectively. Though the competing methods adopt the same degradation model for training, our DMDNet shows better robustness in generating richer and more realistic details, and meanwhile with fewer artifacts. \lxm{Moreover, when the specific references are not available, Ours$^g$ can also perform favorably even though the degradation is unknown, indicating its practical value and flexibility in real-world applications.}
	
	\subsection{Face Identity Recognition}
	\lxmrb{To explore the effectiveness of specific dictionary that brings to the preservation of identity-related details, we adopt Arcface~\cite{deng2019arcface} on our CelebRef-HQ test set and compute the cosine distance between the restored face region and their corresponding ground-truth. 
		From Tables~\ref{tab:generic} and~\ref{tab:specific} one can see that by incorporating the specific dictionary, Ours~(\textit{Full}) has an obvious improvement in Id metric (\eg, at least 8.6\% in $\times 4$ and 30.2\% in $\times 8$ compared with these generic restoration methods). Besides, by using the reference of the same identity, these specific restoration methods perform obviously better than these without identity-related reference. This indicates that 1) these general restoration methods are prone to lose the identity features which is mainly caused by the unrecognizable structures in the severely degraded input, and 2) the specific dictionary is necessary in face restoration task, especially for generating the identity-related textures. Thus, the specific dictionary is essential for preserving higher fidelity and could compensate for the generic dictionary by providing accurate identity features.}


	\begin{figure*}[t]
		\setlength{\abovecaptionskip}{5pt}
		\centering
		\includegraphics[width=\textwidth]{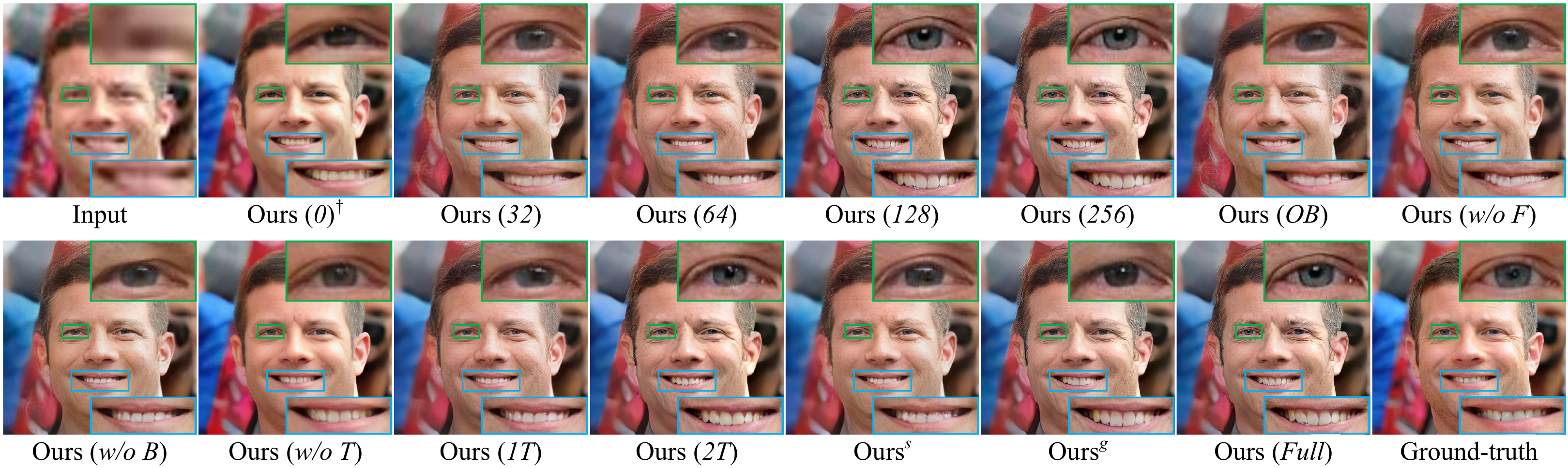}
		\caption{Visual comparison of four groups of DMDNet variants. Best view it by zooming in the screen to see the details.}
		\label{fig:aba}
		\vspace{-2pt}
	\end{figure*}
	
	\subsection{Analyses and Discussions}

	\lxm{Following the former experimental settings, we adopt the test set of CelebRef-HQ on $\times4$ and $\times8$ tasks to analyze and discuss the effectiveness of our DMDNet. 
		We mainly consider four groups of ablation experiments to analyze 1) the item number in the generic dictionary, 2) the optimization manner of the generic dictionary, 3) the multi-scale dictionary transform module, and 4) the necessity of the generic and specific dictionaries. The comparison results are shown in Table~\ref{tab:ablation} and Figure~\ref{fig:aba}, respectively.
		For simplifying the definition, Ours$^s$, Ours$^g$, and Ours (\textit{Full}) denote our DMDNet with the only specific dictionary, our DMDNet with the only generic dictionary, and our DMDNet with the dual dictionaries.}

	\noindent\textbf{(1) \lxm{Item number in the generic dictionary.}} 
	{To explore the item number {$Y$}, we mainly consider five variants, \ie, Ours ($Y$), $Y \in \{0,32,64,128,256\}$, among which Ours (\textit{0}) represents the plain CNNs by directly learning the restoration mapping without considering the external memory dictionaries. For fair evaluation, we increase the channels of the whole framework to 2 times which is denoted as Ours (\textit{0})$^\dagger$. We can see that 1) plain CNNs cannot well learn the blind restoration mappings due to the complex degradation, which may result in worse visual and quantitative results; 2) by increasing the number $Y$, {both the quantitative and visual} performance tend to get better. When $Y=256$, there is only a slight improvement, but it is time- and memory-consuming for training and testing as well as storing such a large dictionary. Thus, in this paper, we adopt Ours (\textit{128}) as the default model.}
	
	\noindent\textbf{(2) Optimization manner of generic dictionary.} 
	\lxm{Since the generation and optimization of the generic dictionary contain three stages, 
		we mainly consider the following three variants,}
	(i) Ours (\textit{OB}): by initializing the generic dictionary with random noise and \textbf{o}nly using \textbf{b}ackward update, (ii) Ours (\textit{w/o F}): by removing the \textbf{f}orward update, (iii) Ours (\textit{w/o B}): by removing the \textbf{b}ackward update. 
	We can see that the performance of Ours (\textit{OB}) is obviously degraded {(1.3 dB in $\times4$ and 1.1 dB in $\times8$)}. We analyze that the long-path way gradient from the final objective loss may optimize the dictionary slowly, resulting in poor results. 
	Compared with Ours (\textit{w/o F}) and Ours (\textit{Full}), the improvement may be mainly ascribed to the usage of forward update.
	Furthermore, by removing the backward update, the PSNR and SSIM of Ours (\textit{w/o B}) have an obvious degradation~(0.6 dB), indicating the effectiveness of backward update in CNNs. 
	\lxm{All these indicate that the initialization, forward and backward update manners can benefit the final restoration results.}

	\begin{table}[t]
		\caption{{Quantitative comparison of four groups of DMDNet variants.}}
		\centering
		\footnotesize
		\renewcommand\arraystretch{1.35}
		\footnotesize
		\setlength{\tabcolsep}{0.06mm}
		{
			\begin{tabular}{|c|c| c c c c| c c c c|}
				\hline
				\rowcolor{lightgray}\rowcolor{lightgray}
				\rowcolor{lightgray}&& \multicolumn{4}{c|}{$\times 4$} & \multicolumn{4}{c|}{$\times 8$} \\
				\rowcolor{lightgray}\multirow{-2}{*}{\makecell[c]{\textbf{Types}}}&\multirow{-2}{*}{\makecell[c]{\textbf{Variants}}}&
				PSNR$\uparrow$ & SSIM$\uparrow$ &  LPIPS$\downarrow$ & FID$\downarrow$ &  PSNR$\uparrow$ & SSIM$\uparrow$ &  LPIPS$\downarrow$ & FID$\downarrow$\\
				\hline \hline
				\multirow{5}{*}{\makecell[c]{\textbf{(1)}}}&Ours (\textit{0})$^\dagger$
				& 26.31  & .832  & .231  & 21.26 & 24.69  & .797  & .292  & 32.44   \\ 
				&Ours (\textit{32})
				& 28.63  & .893  & .180  & 11.01  & 26.65  & .856  & .206  & 14.97   \\ 
				&Ours (\textit{64}) 
				& 28.90  & .900  & .171  & 10.92  & 27.01  & .869  & .199  & 14.63   \\ 
				&Ours (\textit{128})
				& 28.97  & .902  &  .166 & 10.30  & 27.37  & .875  & .194  & 14.43 \\
				&Ours (\textit{256})
				& 28.98  & .902  &  .165 & 10.31  & 27.39  & .876  & .193  & 14.41   \\ 
				\hline
				\multirow{3}{*}{\makecell[c]{\textbf{(2)}}}&Ours (\textit{OB})
				& 27.67  & .880  & .201  & 18.64  & 26.25  & .851  & .240  & 19.29   \\ 
				&Ours (\textit{w/o F}) 
				& 28.60  & .897  & .173  & 14.51  & 27.02  & .869  & .201  & 15.37   \\ 
				&Ours (\textit{w/o B})
				& 28.36  & .896  & .174  & 14.55  & 26.63  & .862  & .205 & 15.46   \\ 
				\hline
				\multirow{3}{*}{\makecell[c]{\textbf{(3)}}}&Ours (\textit{w/o T})
				& 28.24  & .884  & .176  & 13.56  & 26.81  & .863  & .237  & 15.58   \\ 
				&Ours (\textit{1T})
				& 28.83  & .897  & .173  & 12.11  & 27.09  & .870  & .213  & 14.88   \\ 
				&Ours (\textit{2T})
				& 28.90  & .901  & .171  & 10.77  & 27.12  & .871  & .200  & 14.50   \\ 
				\hline
				\multirow{3}{*}{\makecell[c]{\textbf{(4)}}}&Ours$^{s}$
				& 28.43  & .887  & .180  & 13.67  & 26.83  & .860  &  .234 & 15.53   \\ 
				&Ours$^g$
				&{28.91}  & {.901}  & {.169}  & {10.31}  & {27.03}  & {.870}  & {.201}  &  {14.56}   \\
				&Ours (\textit{Full})& \textbf{28.97}  & \textbf{.902}  &  \textbf{.166} & \textbf{10.30}  & \textbf{27.37}  & \textbf{.875}  & \textbf{.194}  & \textbf{14.43} \\
				\hline
		\end{tabular}}
		\label{tab:ablation}
	\end{table}
	
	\noindent\textbf{(3) Multi-scale dictionary transform module.}
	To evaluate the benefits {introduced by multi-scale dictionary transform module}, we conduct three variants, (i) Ours (\textit{w/o T}): by replacing the transform module with the best value that has the highest matching score,  which is similar to ASFFNet~\cite{Li_2020_CVPR}, (ii) Ours (\textit{2T}): by taking {the first two transform modules}, (iii) Ours (\textit{1T}): by taking only {the first one transform module}. {We observe that Ours (\textit{w/o T}) performs on par with ASFFNet* and the individual details are easily lost}. The higher results may be caused by matching both generic and specific dictionaries, but still inferior to Ours (\textit{Full}). By reducing the number of transform modules, the results tend to degrade, \lxm{indicating} the benefits of coarse-to-fine restoration.
	
	\noindent\textbf{(4) Necessity of the generic and specific dictionaries.}
	\lxm{To validate the necessity of the dual dictionaries, we retrain our DMDNet by only using the specific dictionary (Ours$^s$).
		Different from GFRNet~\cite{li2018learning} and ASFFNet~\cite{Li_2020_CVPR}, which adopt only one HQ reference for the final reconstruction, Ours$^s$ utilizes the dictionary transform module by taking all the values into the restoration process. }
	\lxmday{We can see that Ours$^s$ outperforms GFRNet* and ASFFNet*, which indicates that our method can effectively utilize the references for restoration.} Compared with Ours (\textit{Full}), the performance of Ours$^s$ degraded obviously (\eg, mouth). We analyze that this may be caused by the uncertainty of HQ references. When they are not enough to cover all the poses and expressions, Ours$^s$ can obtain limited improvements.
	\lxm{Although Ours$^g$ can generate plausible results, the individual details are easily lost (\eg, pupil color, double-eyelids in Figure~\ref{fig:aba}). 
		So the generic and specific dictionaries can benefit each other.
		By taking the complementary merits of the dual dictionaries in a unified model, Ours (\textit{Full}) achieves the best performance, indicating the necessity of the generic and specific dictionaries.}
	
	\subsection{Model Complexity}
	\lxmrb{Ours~(\textit{Full}) model contains 40.36~\textit{M} parameters (the generic dictionary in \textit{Backward Update} stage has 110.17~\textit{M} parameters but will transfer to model buffer without optimization, which is not included in this value). It takes 122.7 \textit{ms} for Ours~(\textit{Full}) (8 HQ specific reference as example) to restore an image, but can be reduced to 64.8 \textit{ms} by storing the specific dictionary for each identity in advance. The offline stored specific dictionary can be flexibly used in many scenarios, \eg, personal smartphones and famous actors in films. 
		Although the competing methods GFPGAN~\cite{wang2021towards} and GPEN~\cite{Yang2021GPEN} take only 32.2~\textit{ms} on average to handle an image, 
		Ours~(\textit{Full}) performs superior to them in specific restoration, which 
		could restore identity-related textures and preserve better fidelity (see the visual comparison and identity recognition).
	}

	
	\section{Conclusion}\label{sec:5}
	This paper presented a blind face restoration model, \ie, DMDNet, by taking the generic and specific dictionaries in a unified framework. The generic and specific dictionaries store the general and individual facial details, respectively. To handle the degraded input with or without references, dictionary transform module by taking the query-key-value is then suggested to adaptively read and fuse the relevant details from the dual memory dictionaries.
	Finally, multi-scale dictionary transform modules are introduced to benefit the coarse-to-fine restoration. 
	Moreover, we also construct a new face dataset, \ie, CelebRef-HQ, to promote the reference-based restoration methods on handling high-resolution images.
	The experimental results demonstrate the effectiveness of the proposed method in generating photo-realistic results on both synthetic and real-world low-quality images, and show the practical value in different application scenarios.
	
	\ifCLASSOPTIONcompsoc
	\section*{Acknowledgments}
	\else
	\section*{Acknowledgment}
	\fi
	
	This work is partly supported
	by the National Key R\&D Program of China under Grant No. 2021ZD0112100, the~National Natural Science Foundation of China under Grant No. U19A2073, and the Hong Kong RGC RIF grant (R5001-18).
	
	\bibliographystyle{IEEEtran}
	\bibliography{IEEEabrv,egbib}
	\ifCLASSOPTIONcaptionsoff
	\newpage
	\fi

	%

\end{document}